# Comparative Study of ECO and CFNet Trackers in Noisy Environment

Mustansar Fiaz
School of Computer Science and Engineering
Kyungpook National University, Republic of Korea
mustansar@vr.knu.ac.kr

Sajid Javed
Department of Computer Science
University of Warwick, United Kingdom
s.javed.1@warwick.ac.uk

Arif Mahmood
Department of Computer Science and Engineering
Qatar University, Doha, Qatar
arif.mahmood@qu.edu.qa

Soon Ki Jung*
School of Computer Science and Engineering
Kyungpook National University, Republic of Korea
skjung@knu.ac.kr

*Abstract*—**Object tracking is one of the most challenging task and has secured significant attention of computer vision researchers in the past two decades. Recent deep learning based trackers have shown good performance on various tracking challenges. A tracking method should track objects in sequential frames accurately in challenges such as deformation, low resolution, occlusion, scale and light variations. Most trackers achieve good performance on specific challenges instead of all tracking problems, hence there is a lack of general purpose tracking algorithms that can perform well in all conditions. Moreover, performance of tracking techniques has not been evaluated in noisy environments. Visual object tracking has real world applications and there is good chance that noise may get added during image acquisition in surveillance cameras. We aim to study the robustness of two state of the art trackers in the presence of noise including Efficient Convolutional Operators (ECO) and Correlation Filter Network (CFNet). Our study demonstrates that the performance of these trackers degrades as the noise level increases, which demonstrate the need to design more robust tracking algorithms.**

## I. Introduction

Recently, visual object tracking (VOT) has been famous computer vision problem among researchers due to its vast practical applications such as surveillance videos, robotics, human machine interaction, activity recognition and autonomous vehicles [1-5]. Object tracking is an attractive problem for global researchers due to its various challenges such as background clutter, illumination, occlusion, deformation, motion blur, fast motion, in and out planer rotation, out of view, low resolution and scale variation as reported in object tacking benchmark (OTB) 2015 [6]. VOT is defined as the estimation of target location in sequential images in sequence provided the initial position of target in the first frame. Recently, global research community is contributing positively and accomplished very good performances of trackers. Although good performance of tracking algorithms have been attained but there is still an open challenge to design a robust tracking algorithm to handle all OTB challenges even in noisy environment. In physical world, during the image acquisition of surveillance camera, there is a possibility that noise may be added. An ideal tracking algorithm should handle noise efficiently and precisely. Considering real world noise problem, we have evaluated the performance two tracking schemes in the existence of additive white Gaussian noise. In this paper, we have studied the robustness of tracking algorithms systematically and experimentally. We have computed the precision and success plots for single tracking object tracking methods in the presence of additive white Gaussian noise. For experimental evaluation, we have selected object tracking benchmark 2015 [6] which have 100 different videos and target object does not egress from the sequences. Investigation has been performed on six different datasets for each tracker with zero mean and varying variance white additive Gaussian noise.

In literature, several visual object tracking techniques such as classical and contemporary have been discussed by Ali et al. [7] along with VOT applications, challenges, evaluation methods and deliberated annotated datasets. Yilmaz et al. [8] exploited various general purpose trackers, feature representations and data association in his study. Object tracking techniques are categorized into discriminative and generative models by Qin et al. [9]. Qi et al. [10] performed a survey on single object trackers for online learning. Shengping in et al. [10] debated the benefits of trackers based on sparse coding and ordered sparse trackers into appearance model and target search representation trackers. Chen et al. [11] studied trackers based on the Kalman filters in robotic vision. Poppe in 2006 [12] and Jia et al. in 2007 [13] discussed human motion based trackers. In past, each survey is focusing a specific goal and application. Recently, Brekhna et al. [14] have studied the robustness of superpixel algorithms against common types of noises, however no such study has been done for tracking algorithms. Thus robustness of trackers should be evaluated against noise and there exists no such study yet.

## II. Correlation Filter Based Tracking Framework

Recently correlation filter based trackers have attained much attention for object tracking. Correlation filters learn target template in a discriminative way to distinguish object and their translations. Fig. 1 shows the framework of general correlation filter based tracking schemes. Correlation filter based trackers are trained on target on first frame. Target patch is cropped on provided target position on initial frame. For every input frame after initialization, a patch is cropped at predicted location based on previously estimated target position. Feature map is computed from the cropped patch for better input description. To smooth the discontinuities at boundary window, cosine window is convolved with input feature map. Learned correlation filter is convolved with input feature map in Fourier domain resulting response map. Confidence score map is obtained by taking inverse Fourier transform. Maximum value on confidence score map represents the newly estimated target location. Newly estimated target appearance is updated by extracting features at estimated position on current frame and correlation filter is learned with desired output. For extensive experimental results, correlation filter based trackers are selected.

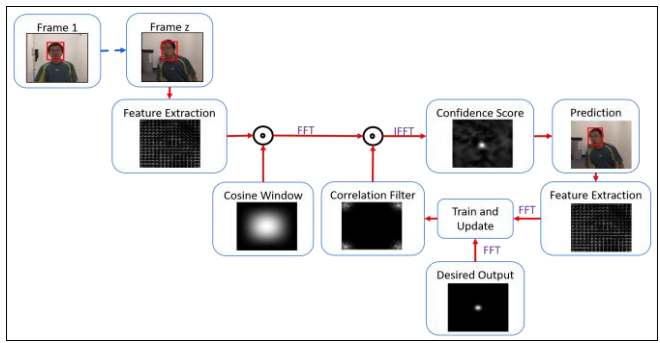

Figure 1: Correlation filter based tracking method. The symbols Θ presents element-wise multiplication, FFT represents the Fast Fourier Transform and IFFT means Inverse Fast Fourier Transform.

## III. Selected Tracking Techniques

### A. ECO: Efficient Convolutional Operators for Tracking

Efficient Convolution Operators (ECO) [15] is based on discriminative correlation filter (DCF) tracker with specialized compact generative model which produce diverse sample space and factorized convolutional operators to reduce number of model parameters. Contrary to other DCF trackers, ECO does not update model on every frame but update after every Nth frame and model is refined using fix number of Conjugate Gradient iterations.

In ECO, factorized convolution operator aimed to reduce the number of feature space. High dimensional feature map is multiplied with factorized convolutional operators resulting reduced feature dimensions. During the tracking of ECO, regression error is minimized in a discriminative way by learning factorized operators and feature coefficient matrix jointly. Loss is optimized using Gauss Newton method [16] and Conjugate Gradient method in frequency domain.

ECO employs compact generative model to produce efficient space by producing reduced number of samples. During the tracking, newly target appearances are represented by Gaussian Mixture Model (GMM) components. A new GMM component is set when a new target appearance appeared during tracking. GMM components are updated using Declercq and Piater online algorithm [17]. GMM components are simplified whenever GMM components exceed threshold. Two close GMM components are merged into a common component if weights are greater than a predefined threshold value else discarded.

### B. End to End Representation Learning For Correlation Filter Based Tracking

In end to end representation learning for correlation filter based tracking [18], is a Siamese based tracking algorithm where similarity is determined whether identical objects are present or not in two image patches. Siamese networks are Y-shaped where two branches integrate into one output layer. Authors used correlation filters (CF) to discriminate image patch from background patches. Author combined CF over CNN efficiently as a differentiable CNN layer by integrating CF as Correlation Filter Network (CFNet) layer. Online tracking in CFNet is performed in forward mode. CFNet back propagates gradient during online learning for optimization of underlying feature representation. Image search space is cropped larger size than previously estimated target size on current frame during CFNet tracking. CNN features are computed for better presentations of inputs. Template features are further given to correlation filters. Template features and search patch features are compared to estimate new target position. Similarity map is obtained by convolving input search features and target template features in Fourier domain. New target location is estimated at maximum score of similarity map. New target appearance is updated on initial target template in a moving average.

## IV. Experiment and Analysis

In this section, a comprehensive experimental analysis of our study is described. We evaluated the tracking schemes on object tracking benchmark OTB 2015 [6]. All experiment have been performed on Processor: Intel(R) Core(TM) i5-4670CPU @ 3.40GHz, RAM: 8:00GB, System Type: 64-bit OS, GPU: GTX 680.

### A. Dataset Formation

For this study, we used OTB 2015 benchmark dataset which consists of 58,879 frames that contains all the twelve object tracking challenges [6]. We used six datasets, one with zero noise and other five contains noise. We have generated five datasets from OTB 2015 by adding white Gaussian noise with different variance. Five sets have additive white Gaussian noise with zero means and varying variance. Variance for each set is defined with the help of

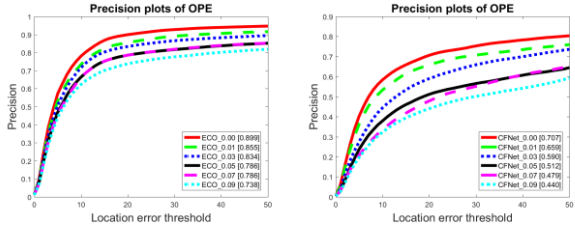

*Figure 2:* Distance precision plots for ECO [15] and CFNet [18] over OTB2015 benchmark [6] using one-pass evaluation (OPE) with additive white Gaussian noise with zero mean and varying variance. The legend contains score at a threshold of 20 pixels for each tracker.

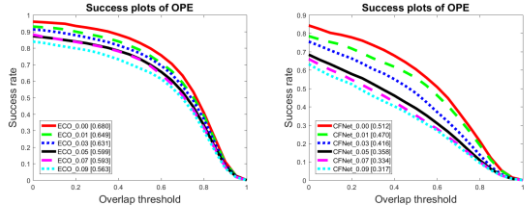

*Figure 3:* Overlap success plots for ECO [15] and CFNet [18] over OTB2015 benchmark [6] using one-pass evaluation (OPE) with additive white Gaussian noise with zero mean and varying variance. The legend contains area under the curve score for each tracker.

arithmetic progression as:

$$a_n = a_1 + (n-1)d \qquad (1)$$

Here $a_n$ defines the $n$ dataset with $n$th variance, $a_1$ is the initial variance is 0.01 while $d$ is 0.02. Qualitative and quantitative analysis have been performed on all datasets.

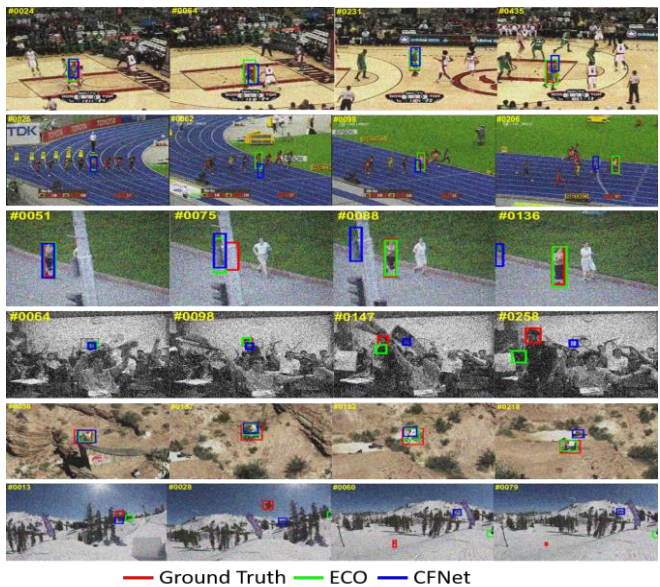

*Figure 4:* Qualitative analysis of trackers ECO [15] and CFNet [18] on OTB2015 [6] containing additive Gaussian noise with zero mean and 0.05 variance on different challenging sequences (from top to bottom *Basketball, Bolt, Jogging-1, Freeman4, MountainBike,* and *Skiing* respectively).

## B. Evaluation Methods

One pass evaluation (OPE) method has been adopted for the evaluation of trackers in the presence of noise. In OPE,

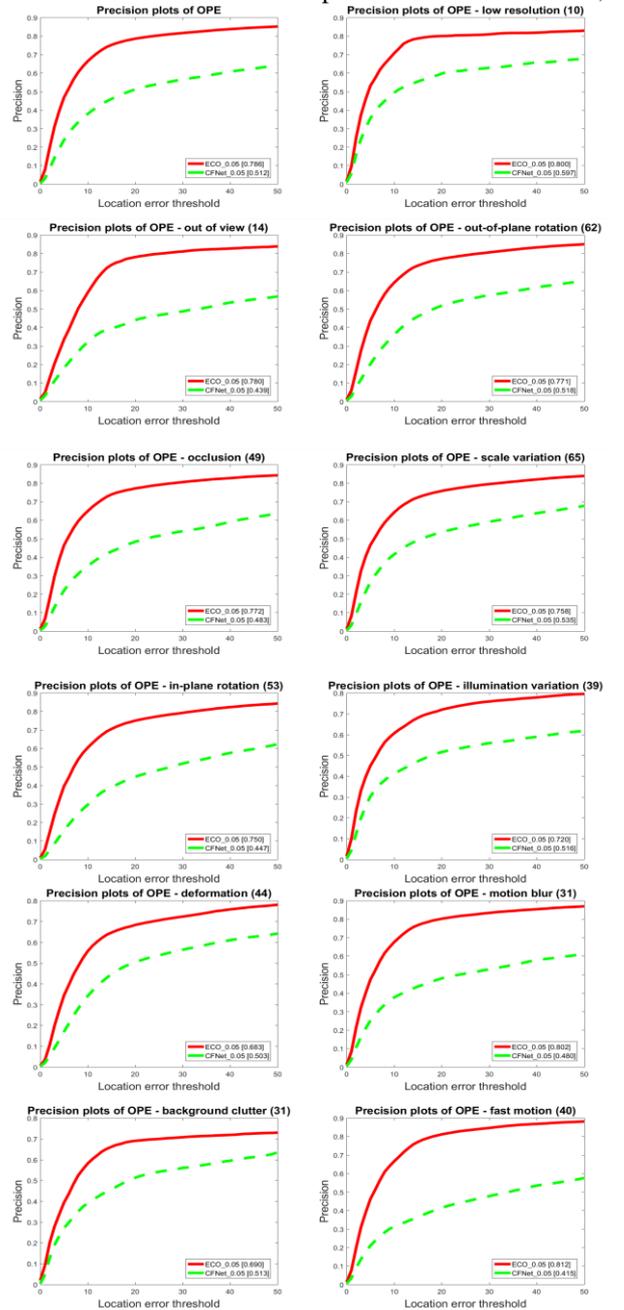

Figure 5: Precision distance plot of trackers ECO and CFNet on OTB2015 containing additive Gaussian noise with zero mean and 0.05 variance over ten different challenges (low resolution, out of view, out of plane rotation, occlusion, scale variation, in-plane rotation, illumination variation, deformation, motion blur, background clutter, and fast motion). The legend contains score at a threshold of 20 pixels for each tracker.

equation on the sequence is performed only once by the tracker. We have computed precision and success for performance of tracking schemes in the existence of white additive noise. To compute the precision, Euclidean distance is calculated between ground-truth center and predicted center of object during tracking. Precision is the percentage of images whose predicted centers lay within the provided threshold from the ground truth centers. Overlap score (OS) is used to compute success plot. Assume, target bounding box is represented by $r_t$ and $r_a$ is used to present the ground-truth. An overlap score is defined as:

$$OS = \frac{|r_t \cap r_a|}{|r_t \cup r_a|} \qquad (2)$$

Here intersection and union of two regions are represented by $\cap$ and $\cap$ respectively, while number of pixels is counted in the area by $|\ |$. Overlap score represents the success rate of the frame in sequence. IF OS score is greater than a threshold, then that frame is referred as successful frame. Success plots shows the success rate of frames in [0 1].

## C. Quantitative Evaluation

Fig. 2 and 3 describes the overall precision and success of the tracking algorithms with and without presence of noise. From the plots, our investigation clearly says that ECO tracking method has very less impact of noise. On the contrary to CFNet performance degrades rapidly as the noise increases. ECO tracker shows almost same response with noise of 0.05 and 0.07. We noticed that ECO tracking algorithm showed better performance compared to CFNet which performed poorly even in the presence of additive white Gaussian noise with varying variance.

Fig. 5 shows the precision performance of ECO and CFNet trackers over 11 tacking challenges. We have computed plots over dataset having additive white Gaussian noise with zero mean and variance of 0.05. Fig. 5 shows that ECO better than CFNet on every object tracking challenge.

## D. Qualitative Evaluation

Qualitative study of tracking algorithms have been performed and shown in fig. 4. Fig. 4 shows tracking results for randomly selected sequences from OTB-100 dataset with additive white Gaussian noise of zero mean and 0.05 variance. Sequences are selected randomly covering all the occlusion, deformation, illumination, fast motion and motion blur tracking challenges. ECO is showing better results in *Basketball*, *Bolt*, *Jogging-1*, and *MountainBike* sequences. Both trackers are exhibiting poor results over *Freeman4* and *Skiing* sequences. By analyzing fig. 3, we observed that ECO performed better than CFNet tracking algorithm.

## V. Conclusion

In this study, we examined the robustness of ECO and CFNet trackers over additive white Gaussian noise. A comprehensive evaluation has been performed for those trackers in the existence of additive white Gaussian noise. For experimental investigation OTB100 benchmark has been used. Precision and success plots for trackers with varying variance of Gaussian noise are computed. Our study shows that ECO is less sensitive to noise comparing with CFNet tracker. ECO performed better than CFNet even in the presence of noise. In future, we aim to include more tracking algorithms with different types of noises and will provide comprehensive study of the robustness of trackers to noises.

## Acknowledgment

This research was supported by Development project of leading technology for future vehicle of the business of Daegu metropolitan city (No. 20171105).